\documentclass[11pt,a4paper]{article}
\usepackage[hyphens]{url}
\usepackage[hyperref]{acl2017}
\usepackage{times}
\usepackage{latexsym}

\usepackage[utf8]{inputenc} 
\usepackage[T1]{fontenc}    

\usepackage{booktabs}       
\usepackage{amsfonts}       
\usepackage{nicefrac}       
\usepackage{microtype}      

\usepackage{amsmath}
\usepackage{verbatim}
\usepackage{amssymb}
\usepackage{amsthm}
\usepackage{color}
\usepackage{graphicx}
\usepackage{subfigure}

\aclfinalcopy 
\def\aclpaperid{547} 

\newcommand{\ignore}[1]{}

\DeclareSymbolFont{extraup}{U}{zavm}{m}{n}
\DeclareMathSymbol{\heartsuit}{\mathalpha}{extraup}{86}
\DeclareMathSymbol{\diamondsuit}{\mathalpha}{extraup}{87}

\title{\ignore{ Neural Cache Language Model\\}
       \ignore{Learning to Discover and Reuse Words in Open Vocabulary Language Modeling\\}
       \ignore{Memorization and Reuse in Open Vocabulary Neural Language Modeling}
       \ignore{Open Vocabulary Language Modeling with a Continuous Cache}
       Learning to Create and Reuse Words in \\ Open-Vocabulary Neural Language Modeling
       }
\author{Kazuya Kawakami$^{\spadesuit}$ \quad
		Chris Dyer$^{\clubsuit}$
		\quad Phil Blunsom$^{\spadesuit\clubsuit}$\\
$^{\spadesuit}$Department of Computer Science, University of Oxford, Oxford, UK \\
$^{\clubsuit}$DeepMind, London, UK \\
{\tt \{kazuya.kawakami,phil.blunsom\}@cs.ox.ac.uk,cdyer@google.com}
}

\newcommand{\unk}{\langle \textsc{unk} \rangle}
\newcommand{\pb}[1]{\textcolor{red}{\bf\small [#1 --PB]}}
\newcommand{\cd}[1]{\textcolor{blue}{\bf\small [#1 --CD]}}
\newcommand{\kk}[1]{\textcolor{magenta}{\bf\small [#1 --KK]}}

\date{}

\begin{document}

\maketitle

\begin{abstract}
Fixed-vocabulary language models fail to account for one of the most characteristic statistical facts of natural language: the frequent creation and reuse of new word types. Although character-level language models offer a partial solution in that they can create word types not attested in the training corpus, they do not capture the ``bursty'' distribution of such words. In this paper, we augment a hierarchical $\mathrm{LSTM}$  language model that generates sequences of word tokens character by character with a caching mechanism that learns to reuse previously generated words. 
To validate our model we construct a new open-vocabulary language modeling corpus (the Multilingual Wikipedia Corpus; MWC) from comparable Wikipedia articles in 7 typologically diverse languages and demonstrate the effectiveness of our model across this range of languages.
\end{abstract}

\section{Introduction}\label{sec:intro}
\ignore{
\cd{analysis idea: sample from a character LM, and give statistics about new words and their reuse. Sample from the hierarchical char LM with cache and show their statistics. We should see something that looks more like the empirical distribution. I bet newly created words are never reused. Also, another baseline we might want to run is a PYP LM. I've got code for this.}
\kk{I haven't implemented decoder though ... How about visualizing $\lambda$?}\cd{yes, that would be useful too. Anyway, all of this is low priority until we have numbers!}
}
\ignore{I've got to run now, but here's my high level message on the intro. You need to motivate this paper by saying that we know things about how language is used. There are two empirical facts we know. First, Heaps (1978) showed that vocabularies keep growing. Second, Church (2000) showed that rare words are ``bursty''. Character models solve the first challenge, but we need a cache to solve the second. While there are cache models, it's especially important that we put both of these things together. It's worth point out that hierarchical Bayesian language models (e.g. Goldwater, Griffiths, Johnson JMLR 2011; Chahuneau et al. NAACL, 2014) do solve this, but they don't enjoy the benefits of unbounded dependencies, distributed representations, etc. 

In this paper, we create a model that has all three of these things: new words, burstiness, and long-range dependencies based on RNNs.}

Language modeling is an important problem in natural language processing with many practical applications (translation, speech recognition, spelling autocorrection, etc.). Recent advances in neural networks provide strong representational power to language models with distributed representations and unbounded dependencies based on recurrent networks (RNNs). However, most  language models operate by generating words by sampling from a closed vocabulary which is composed of the most frequent words in a corpus. Rare tokens are typically replaced by a special token, called the unknown word token, $\unk$. Although  fixed-vocabulary language models have some important practical applications and are appealing models for study, they fail to capture two empirical facts about the distribution of words in natural languages. First, vocabularies keep growing as the number of documents in a corpus grows: new words are constantly being created~\cite{heaps1978information}. Second, rare and newly created words often occur in ``bursts'', i.e., once a new or rare word has been used once in a document, it is often repeated~\cite{church:1995,church2000empirical}.

The open-vocabulary problem can be solved by dispensing with word-level models in favor of models that predict sentences as sequences of characters~\cite{sutskever2011generating,chung2016hierarchical}. Character-based models are quite successful at learning what (new) word forms look like (e.g., they learn a language's orthographic conventions that tell us that \emph{sustinated} is a plausible English word and \emph{bzoxqir} is not) and, when based on models that learn long-range dependencies such as RNNs, they can also be good models of how words fit together to form sentences.

However, existing character-sequence models have no explicit mechanism for modeling the fact that once a rare word is used, it is likely to be used again. In this paper, we propose an extension to character-level language models that enables them to reuse previously generated tokens~(\S\ref{sec:model}). Our starting point is a hierarchical $\mathrm{LSTM}$  that has been previously used for modeling sentences (word by word) in a conversation~\cite{sordoni2015hierarchical}, except here we model words (character by character) in a sentence. To this model, we add a caching mechanism similar to recent proposals for caching that have been advocated for closed-vocabulary models~\cite{merity2016pointer,grave2016improving}. As word tokens are generated, they are placed in an LRU cache, and, at each time step the model decides whether to copy a previously generated word from the cache or to generate it from scratch, character by character. The decision of whether to use the cache or not is a latent variable that is marginalised during learning and inference. In summary, our model has three properties: it creates new words, it accounts for their burstiness using a cache, and, being based on $\mathrm{LSTM}$ s over word representations, it can model long range dependencies.

To evaluate our model, we perform ablation experiments with variants of our model without the cache or hierarchical structure. In addition to standard English data sets (PTB and WikiText-2), we introduce a new multilingual data set: the Multilingual Wikipedia Corpus (MWC), which is constructed from comparable articles from Wikipedia in 7 typologically diverse languages (\S\ref{sec:dataset}) and show the effectiveness of our model in all languages (\S\ref{sec:experiments}). By looking at the posterior probabilities of the generation mechanism (language model vs. cache) on held-out data, we find that the cache is used to generate ``bursty'' word types such as proper names, while numbers and generic content words are generated preferentially from the language model~(\S\ref{sec:analysis}).

\section{Model}\label{sec:model}
In this section, we describe our hierarchical character language model with a word cache. As is typical for RNN language models, our model uses the chain rule to decompose the problem into incremental predictions of the next word conditioned on the history:
\begin{align*}
    p(\boldsymbol{w}) = \prod_{t=1}^{|\boldsymbol{w}|} p(w_{t} \mid \boldsymbol{w}_{<t}).
\end{align*}

We make two modifications to the traditional RNN language model, which we describe in turn. First, we begin with  a cache-less model we call the hierarchical character language model ($\mathrm{HCLM}$; \S\ref{sec:hclm}) which generates words as a sequence of characters and constructs a ``word embedding'' by encoding a character sequence with an $\mathrm{LSTM}$ ~\cite{ling2015finding}. However, like conventional closed-vocabulary, word-based models, it is based on an $\mathrm{LSTM}$  that conditions on words represented by fixed-length vectors.\footnote{The $\mathrm{HCLM}$ is an adaptation of the hierarchical recurrent encoder-decoder of \newcite{sordoni2015hierarchical} which was used to model dialog as a sequence of actions sentences which are themselves sequences of words. The original model was proposed to compose words into query sequences but we use it to compose characters into word sequences.}

The $\mathrm{HCLM}$ has no mechanism to reuse words that it has previously generated, so new forms will only be repeated with very low probability. However, since the $\mathrm{HCLM}$ is not merely generating sentences as a sequence of characters, but also segmenting them into words, we may add a word-based cache to which we add words keyed by the hidden state being used to generate them (\S\ref{sec:cache}). This cache mechanism is similar to the model proposed by \newcite{merity2016pointer}.

\paragraph{Notation.}
Our model assigns probabilities to sequences of words $\boldsymbol{w} = w_{1}, \ldots, w_{|\boldsymbol{w}|}$, where $|\boldsymbol{w}|$ is the length, and where each word $w_{i}$ is represented by a sequence of characters $\boldsymbol{c}_{i} = c_{i,1}, \ldots, c_{i,{|\boldsymbol{c_i}|}}$ of length $|\boldsymbol{c}_i|$. 

\subsection{Hierarchical Character-level Language Model ($\mathrm{HCLM}$)}\label{sec:hclm}

This hierarchical model satisfies our linguistic intuition that written language has (at least) two different units, characters and words. 

\begin{figure*}[htbp]
    \begin{center}
        \includegraphics[width=1.0\linewidth]{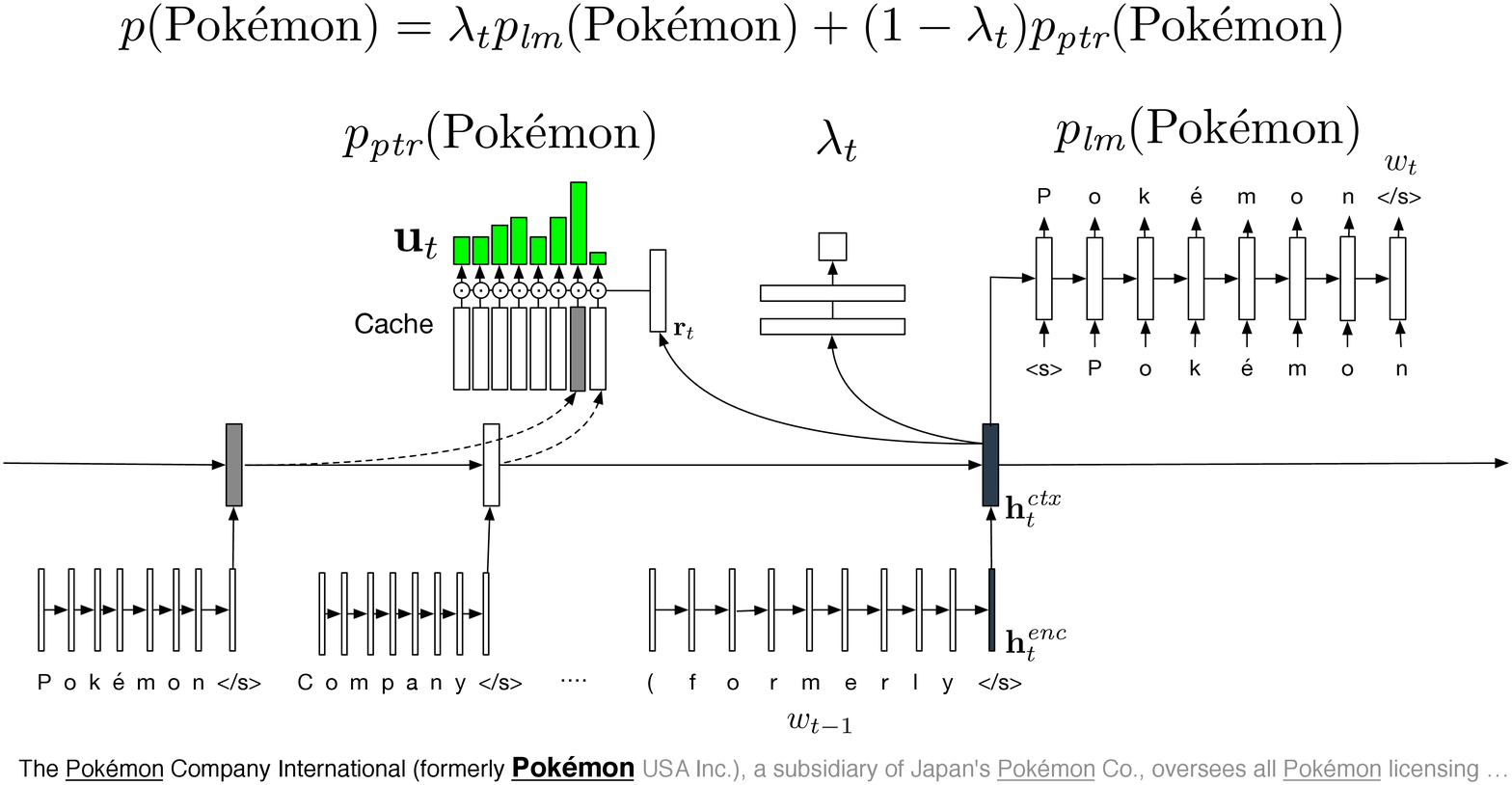}
    \end{center}
    \caption{Description of Hierarchical Character Language Model with Cache.}
    \label{fig:model}
\end{figure*}

The $\mathrm{HCLM}$ consists of four components, three $\mathrm{LSTM}$s~\cite{hochreiter1997long}: a character encoder, a word-level context encoder, and a character decoder (denoted $\mathrm{LSTM}_{\textit{enc}}$, $\mathrm{LSTM}_{\textit{ctx}}$, and $\mathrm{LSTM}_{\textit{dec}}$, respectively), and a softmax output layer over the character vocabulary. Fig.~\ref{fig:model} illustrates an unrolled $\mathrm{HCLM}$.

Suppose the model reads word $w_{t-1}$ and predicts the next word $w_{t}$. First, the model reads the character sequence representing the word $w_{t-1} = c_{t-1,1}, \ldots, c_{t-1,|\boldsymbol{c}_{t-1}|}$ where $|\boldsymbol{c}_{t-1}|$ is the length of the word generated at time $t-1$ in characters. Each character is represented as a vector $\mathbf{v}_{c_{t-1,1}}, \ldots, \mathbf{v}_{c_{t-1,|\boldsymbol{c}_{t-1}|}}$ and fed into the encoder $\mathrm{LSTM}_{\textit{enc}}$ . The final hidden state of the encoder $\mathrm{LSTM}_{\textit{enc}}$ is used as the vector representation of the previously generated word $w_{t-1}$,
\begin{align*}
	\mathbf{h}_{t}^{\textit{enc}} = \mathrm{LSTM}_{\textit{enc}}(\mathbf{v}_{c_{t-1,1}}, \ldots, \mathbf{v}_{c_{t-1,|\boldsymbol{c}_t|}}).
\end{align*}

Then all the vector representations of words $(\mathbf{v}_{w_{1}}, \ldots, \mathbf{v}_{w_{|\boldsymbol{w}|}})$ are processed with a context $\mathrm{LSTM}_{\textit{ctx}}$ . Each of the hidden states of the context $\mathrm{LSTM}_{\textit{ctx}}$  are considered representations of the history of the word sequence.
\begin{align*}
	\mathbf{h}_{t}^{\textit{ctx}} = \mathrm{LSTM}_{\textit{ctx}}(
	\mathbf{h}_{1}^{\textit{enc}}, \ldots, \mathbf{h}_{t}^{\textit{enc}})
\end{align*} 

Finally, the initial state of the decoder $\mathrm{LSTM}$  is set to be $\mathbf{h}_{t}^{\textit{ctx}}$ and the decoder $\mathrm{LSTM}$  reads a vector representation of the start symbol $\mathbf{v}_{\langle \textsc{s} \rangle}$ and generates the next word $w_{t+1}$ character by character.\ignore{\cd{Is $\mathbf{h}_{t}$ used to initialize the decoder $\mathrm{LSTM}$ 's $\mathbf{c}_0$?}\kk{$h_{w_{t}}$ is used to initialize $\mathbf{h}_{0}$ of the decoder $\mathrm{LSTM}$ .}\cd{it's not immediately clear what $j$ is here. Also, since this is a different $\mathbf{h}$, you might have a superscript $\mathbf{h}^{\textit{dec}}_{t+1,j}$ or something.}} To predict the $j$-th character in $w_{t}$, the decoder $\mathrm{LSTM}$ reads vector representations of the previous characters in the word, conditioned on the context vector $\mathbf{h}_{t}^{
\textit{ctx}}$ and a start symbol.
\begin{align*}
	\mathbf{h}_{t,j}^{\textit{dec}} = \mathrm{LSTM}_{\textit{dec}}(\mathbf{v}_{c_{t,1}}, \ldots, \mathbf{v}_{c_{t,j-1}}, \mathbf{h}_{t}^{\textit{ctx}}, \mathbf{v}_{\langle \textsc{s} \rangle}).
\end{align*}

The character generation probability is defined by a softmax layer for the corresponding hidden representation of the decoder $\mathrm{LSTM}$ .
\begin{align*}
	p(c_{t,j}\mid \boldsymbol{w}_{<t}, \boldsymbol{c}_{t,<j}) = \mathrm{softmax}(\mathbf{W}_{\textit{dec}}\mathbf{h}_{t,j}^{\textit{dec}} + \mathbf{b}_{\textit{dec}})	
\end{align*}

\ignore{
\cd{use mathrm{softmax} instead of just softmax}.\cd{for structured objects like sequences of words, use boldsymbol{w} instead of just w. Since $w_{<t}$ is a sequence of words, it should be $\boldsymbol{w}_{<t}$.}
}

Thus, a word generation probability from $\mathrm{HCLM}$ is defined as follows.
\begin{align*}
	p_{lm}(w_{t}\mid \boldsymbol{w}_{<t}) = \prod_{j=1}^{|\boldsymbol{c}_t|} p(c_{t,j}\mid \boldsymbol{w}_{<t}, \boldsymbol{c}_{t,<j})
\end{align*}

\ignore{
\cd{There's something weird about predicting $w_t$ in terms of $c_{t+1,\cdot}$. I think we should settle on predicting $w_t$ and $c_{t,\cdot}$, always conditioning on $\boldsymbol{x}_{<t}$ (i.e., I don't want to see any $t+1$, or really any $t-1$---if you need to refer to the past, use $foo_{<t}$. Think about $t$ as being answering the question ``what am I predicting right now?''.  Then $\mathbf{h}_t$ always means ``all the information I have to make the prediction at time $t$.}\kk{What should I do in the previous section? I have to use a lot of $t-1$ if we predict $w_{t}$...}\kk{Looks good now?}\cd{I made one change to call it $\mathbf{h}_{t}^{\textit{enc}}$ rather than $\mathbf{h}_{t-1}^{\textit{enc}}$, and a few other changes to get rid of $M$ and $N$ which look like constants but aren't.}\kk{Representation of word $w_{t-1}$ is $\mathbf{h}_{t}$? Isn't it confusing?}
}

\ignore{
\cd{for probability p, use $p$ if you're talking about your model. If you are talking about an event space, use $P$, if you are talking about probabilities in theory (independent of any modeling decisions), $\Pr$ might be appropriate.}
}

\subsection{Continuous cache component}
\label{sec:cache}
\ignore{
\kk{TODO: $i$ and $t$ is not used consistently across sections. $i$ is used to denote word index, $j$ denotes character index in a word $w_{i}$. Here $i$ is used to denote memory index. Maybe I should use $t$ for word index, but what is the best subscript of character?}\cd{There seem to be three indices we need: word index (measured in terms of tokens). Let's call that $t$. Then there is the character index inside words. Let's call that $j$. Then there is the index into a memory slot, lets call that $i$. Sound ok?}\kk{Fixed. TODO: Check. Fix figure}
}

The cache component is an external memory structure which store $K$ elements of recent history. Similarly to the memory structure used in \newcite{grave2016improving}, a word is added to a key-value memory after each generation of $w_t$. The key at position $i \in [1,K]$ is $\mathbf{k}_i$ and its value $m_i$. The memory slot is chosen as follows: if the $w_t$ exists already in the memory, its key is updated (discussed below). Otherwise, if the memory is not full, an empty slot is chosen or the least recently used slot is overwritten. When writing a new word to memory, the key is the RNN representation that was used to generate the word ($\mathbf{h}_t$) and the value is the word itself ($w_t$). In the case when the word already exists in the cache at some position $i$, the $\mathbf{k}_i$ is updated to be the arithmetic average of $\mathbf{h}_t$ and the existing $\mathbf{k}_i$.

To define the copy probability from the cache at time $t$, a distribution over copy sites is defined using the attention mechanism of \newcite{bahdanau2014neural}. To do so, we construct a query vector ($\mathbf{r}_t$) from the RNN's current hidden state $\mathbf{h}_t$,
\begin{align*}
	\mathbf{r}_{t} = \text{tanh}(\mathbf{W}_{q}\mathbf{h}_{t} + \mathbf{b}_{q}),
\end{align*}
then, for each element $i$ of the cache, a `copy score,' $u_{i,t}$ is computed,
\begin{align*}
	u_{i,t} = \mathbf{v}^{T}\text{tanh}(\mathbf{W}_{u}\mathbf{k}_{i} + \mathbf{r}_{t}).
\end{align*}
Finally, the probability of generating a word via the copying mechanism is:
\begin{align*}
	p_{\textit{mem}}(i \mid \mathbf{h}_{t}) &= \mathrm{softmax}_i(\mathbf{u}_{t}) \\
	p_{\textit{ptr}}(w_t \mid \mathbf{h}_{t}) &= p_{\textit{mem}}(i \mid \mathbf{h}_{t})[m_i = w_t],
\end{align*}
where $[m_i = w_t]$ is 1 if the $i$th value in memory is $w_t$ and 0 otherwise. Since $p_{\textit{mem}}$ defines a distribution of slots in the cache, $p_{\textit{ptr}}$ translates it into word space.

\subsection{Character-level Neural Cache Language Model}
The word probability $p(w_{t}\mid \boldsymbol{w}_{<t})$ is defined as a mixture of the following two probabilities. The first one is a language model probability, $p_{\textit{lm}}(w_{t}\mid \boldsymbol{w}_{<t})$ and the other is pointer probability , $p_{\textit{ptr}}(w_{t}\mid \boldsymbol{w}_{<t})$. The final probability $p(w_{t}\mid \boldsymbol{w}_{<t})$ is 
\begin{align*}
	\lambda_t p_{\textit{lm}}(w_{t}\mid \boldsymbol{w}_{<t}) + (1-\lambda_t)p_{\textit{ptr}}(w_{t}\mid \boldsymbol{w}_{<t}),
\end{align*}
where $\lambda_t$ is computed by a multi-layer perceptron with two non-linear transformations using $\mathbf{h}_t$ as its input, followed by a transformation by the logistic sigmoid function:
\begin{align*}
    \gamma_t = \mathrm{MLP}(\mathbf{h}_t), \qquad \lambda_t = \frac{1}{1-e^{-\gamma_t}}.
\end{align*}
We remark that \newcite{grave2016improving} use a clever trick to estimate the probability, $\lambda_t$ of drawing from the LM by augmenting their (closed) vocabulary with a special symbol indicating that a copy should be used. This enables word types that are highly predictive in context to compete with the probability of a copy event. However, since we are working with an open vocabulary, this strategy is unavailable in our model, so we use the MLP formulation.

\ignore{The experiments show that this adaptive parameter is necessary to obtain better\cd{better relative to what? something where $\lambda_t$ is always 0.5? or always fixed?} performance. \kk{reference}}

\ignore{\subsection{Parameter Learning\pb{We don't need this heading.}}}
\subsection{Training objective}
The model parameters as well as the character projection parameters are jointly trained by maximizing the following log likelihood of the observed characters in the training corpus,
\begin{align*}
	\mathcal{L} = -\sum \log p(w_{t}\mid \boldsymbol{w}_{<t}).
\end{align*}

\section{Datasets}\label{sec:dataset}
We evaluate our model on a range of datasets, employing preexisting benchmarks for comparison to previous published results, and a new multilingual corpus which specifically tests our model's performance across a range of typological settings.

\subsection{Penn Tree Bank (PTB)}
We evaluate our model on the Penn Tree Bank. For fair comparison with previous works, we followed the standard preprocessing method used by~\newcite{mikolov2010recurrent}. In the standard preprocessing, tokenization is applied, words are lower-cased, and punctuation is removed. Also, less frequent words are replaced by unknown an token (UNK),\footnote{When the unknown token is used in character-level model, it is treated as if it were a normal word (i.e. UNK is the sequence U, N, and K). This is somewhat surprising modeling choice, but it has become conventional~\cite{chung2016hierarchical}.} constraining the word vocabulary size to be 10k. Because of this preprocessing, we do not expect this dataset to benefit from the modeling innovations we have introduced in the paper. Fig.\ref{tb:ptb} summarizes the corpus statistics.

\begin{table}[ht]
    \begin{center}
    \begin{tabular}{lrrr}
    \toprule[0.3ex]
          & Train & Dev     & Test      \\
    \midrule[0.2ex]
    Character types & 50 & 50 & 48\\[0.2ex]
    Word types  & 10000 & 6022 & 6049\\[0.2ex]
    OOV rate & - & \textbf{0.00\%} & \textbf{0.00\%}\\[0.2ex]
    Word tokens &  0.9M & 0.1M & 0.1M\\[0.2ex]
    Characters & 5.1M & 0.4M & 0.4M\\[0.2ex]
    \bottomrule[0.3ex]
    \end{tabular}
    \end{center}
        \vskip1ex
        \caption{PTB Corpus Statistics.\label{tb:ptb}}
\end{table}

\subsection{WikiText-2}
\newcite{merity2016pointer} proposed the WikiText-2 Corpus as a new benchmark dataset.\footnote{\url{http://metamind.io/research/the-wikitext-long-term-dependency-language-modeling-dataset/}} They pointed out that the preprocessed PTB is unrealistic for real language use in terms of word distribution. Since the vocabulary size is fixed to 10k, the word frequency does not exhibit a long tail. The wikiText-2 corpus is constructed from 720 articles. They provided two versions. The version for word level language modeling was preprocessed by discarding infrequent words. But, for character-level models, they provided raw documents without any removal of word or character types or lowercasing, but with tokenization. We make one change to this corpus: since Wikipedia articles make extensive use of characters from other languages; we replaced character types that occur fewer than 25 times were replaced with a dummy character (this plays the role of the $\unk$ token in the character vocabulary). Tab.~\ref{tb:wiki2} summarizes the corpus statistics.

\begin{table}[ht]
    \begin{center}
    \begin{tabular}{lrrr}
    \toprule[0.3ex]
                    & Train & Dev    & Test\\
    \midrule[0.2ex]
    Character types & 255   & 128    & 138\\[0.2ex]
    Word types      & 76137 & 19813  & 21109\\[0.2ex]
    OOV rate        & -     & 4.79\% & 5.87\%\\[0.2ex]
    Word tokens     & 2.1M  & 0.2M   & 0.2M\\[0.2ex]
    Characters      & 10.9M & 1.1M   & 1.3M\\[0.2ex]
    \bottomrule[0.3ex]
    \end{tabular}
    \end{center}
        \vskip1ex
        \caption{WikiText-2 Corpus Statistics.\label{tb:wiki2}}
\end{table}

\subsection{Multilingual Wikipedia Corpus (MWC)}
Languages differ in what word formation processes they have. For character-level modeling it is therefore interesting to compare a model's performance across languages. Since there is at present no standard multilingual language modeling dataset, we created a new dataset, the Multilingual Wikipedia Corpus (MWC), a corpus of the same Wikipedia articles in 7 languages which manifest a range of morphological typologies. The MWC contains English (EN), French (FR), Spanish (ES), German (DE), Russian (RU), Czech (CS), and Finnish (FI).

To attempt to control for topic divergences across languages, every language's data consists of the same articles. Although these are only comparable (rather than true translations), this ensures that the corpus has a stable topic profile across languages.\footnote{The Multilingual Wikipedia Corpus (MWC) is available for download from \url{http://k-kawakami.com/research/mwc
}}
\paragraph{Construction \& Preprocessing}
We constructed the MWC similarly to the WikiText-2 corpus. Articles were selected from Wikipedia in the 7 target languages\ignore{ using \cd{what criteria? Just length?}\kk{Yes. No quality annotation.}}. To keep the topic distribution to be approximately the same across the corpora, we extracted articles about entities which explained in all the languages. 
We extracted articles which exist in all languages and each consist of more than 1,000 words, for a total of 797 articles. These cross-lingual articles are, of course, not  usually translations, but they tend to be comparable. This filtering ensures that the topic profile in each language is similar. Each language corpus is approximately the same size as the WikiText-2 corpus.

Wikipedia markup was removed with WikiExtractor,\footnote{\url{https://github.com/attardi/wikiextractor}} to obtain plain text. We used the same thresholds to remove rare characters in the WikiText-2 corpus. No tokenization or other normalization (e.g., lowercasing) was done.

\paragraph{Statistics}
After the preprocessing described above, we randomly sampled 360 articles.\ignore{\pb{Should be consistent in using 'entity' or 'article'}\kk{Articles will be used}} The articles are split into 300, 30, 30 sets and the first 300 articles are used for training and the rest are used for dev and test respectively. Table \ref{tb:uwc} summarizes the corpus statistics.

\begin{table*}[ht]
    \begin{center}\small
    \begin{tabular}{cr@{\hskip 9pt}r@{\hskip 9pt}r@{\hskip 9pt}r@{\hskip 9pt}r@{\hskip 9pt}r@{\hskip 9pt}r@{\hskip 9pt}r@{\hskip 9pt}r@{\hskip 9pt}r@{\hskip 9pt}r@{\hskip 9pt}r@{\hskip 9pt}r@{\hskip 9pt}r@{\hskip 9pt}r}
    \toprule[0.3ex]
      & \multicolumn{3}{c}{Char. Types} & \multicolumn{3}{c}{Word Types} & \multicolumn{2}{c}{OOV rate} &
      \multicolumn{3}{c}{Tokens}  & \multicolumn{3}{c}{Characters}\\
    \midrule[0.2ex]
     & Train & Valid & Test & Train & Valid & Test & Valid & Test & Train & Valid & Test & Train & Valid & Test\\
    \midrule[0.2ex]
    EN & 307 & 160 & 157 & 193808 & 38826 & 35093 & 6.60\% & 5.46\% & 2.5M & 0.2M & 0.2M & 15.6M & 1.5M & 1.3M\\[0.2ex]
    FR & 272 & 141 & 155 & 166354 & 34991 & 38323 & 6.70\% & 6.96\% & 2.0M & 0.2M & 0.2M & 12.4M & 1.3M & 1.6M\\[0.2ex]
    DE & 298 & 162 & 183 & 238703 & 40848 & 41962 & 7.07\% & 7.01\% & 1.9M & 0.2M & 0.2M & 13.6M & 1.2M & 1.3M\\[0.2ex]
    ES & 307 & 164 & 176 & 160574 & 31358 & 34999 & 6.61\% & 7.35\% & 1.8M & 0.2M & 0.2M & 11.0M & 1.0M & 1.3M\\[0.2ex]
    CS & 238 & 128 & 144 & 167886 & 23959 & 29638 & 5.06\% & 6.44\% & 0.9M & 0.1M & 0.1M & 6.1M & 0.4M & 0.5M\\[0.2ex]
    FI & 246 & 123 & 135 & 190595 & 32899 & 31109 & 8.33\% & 7.39\% & 0.7M & 0.1M & 0.1M & 6.4M & 0.7M & 0.6M\\[0.2ex]
    RU & 273 & 184 & 196 & 236834 & 46663 & 44772 & 7.76\% & 7.20\% & 1.3M & 0.1M & 0.1M & 9.3M & 1.0M & 0.9M\\[0.2ex]
    \bottomrule[0.3ex]
    \end{tabular}
    \end{center}
        \vskip-0.2ex
        \caption{Summary of MWC Corpus.
        \label{tb:uwc}}
        \vskip-0.8ex
\end{table*}

\begin{figure}[htbp]
    \centering
        \includegraphics[width=1.0\linewidth]{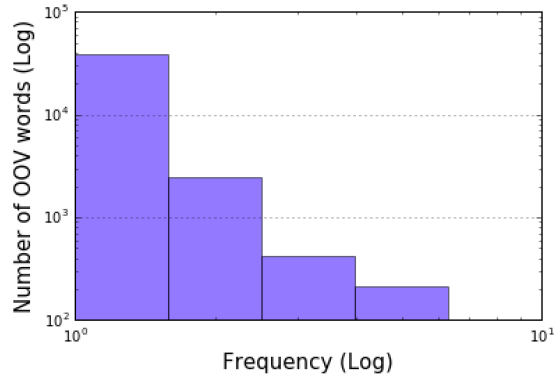}
    \caption{Histogram of OOV word frequencies in the dev$+$test part of the MWC Corpus (EN).}
    \label{fig:freq}
\end{figure}

Additionally, we show in Fig.~\ref{fig:freq} the distribution of frequencies of OOV word types (relative to the training set) in the dev$+$test portions of the corpus, which shows a power-law distribution, which is expected for the burstiness of rare words found in prior work. Curves look similar for all languages (see Appendix~\ref{sec:appendix-corpus}).

\section{Experiments}\label{sec:experiments}
We now turn to a series of experiments to show the value of our hierarchical character-level cache language model. For each dataset we trained the model with $\mathrm{LSTM}$  units. To compare our results with a strong baseline, we also train a model without the cache.

\paragraph{Model Configuration}
For $\mathrm{HCLM}$ and $\mathrm{HCLM}$ with cache models, We used 600 dimensions for the character embeddings and the $\mathrm{LSTM}$s have 600 hidden units for all the experiments. This keeps the model complexity to be approximately the same as previous works which used an $\mathrm{LSTM}$ with 1000 dimension. Our baseline $\mathrm{LSTM}$ have 1000 dimensions for embeddings and reccurence weights.

For the cache model, we used cache size 100 in every experiment. All the parameters including character projection parameters are randomly sampled from uniform distribution from $-0.08$ to $0.08$. The initial hidden and memory state of $\mathrm{LSTM}_{\textit{enc}}$ and $\mathrm{LSTM}_{\textit{ctx}}$ are initialized with zero. 
Mini-batches of size 25 are used for PTB experiments and 10 for WikiText-2, due to memory limitations. The sequences were truncated with 35 words. Then the words are decomposed to characters and fed into the model. A Dropout rate of $0.5$ was used for all but the recurrent connections.

\ignore{
For PTB experiments, we used 600 dimensions for the character embeddings, and the $\mathrm{LSTM}$ s have 600 hidden units. This keeps the model complexity to be approximately the same as previous works which used an $\mathrm{LSTM}$  with 1000 dimension. 
As for WikiText-2 and MWC experiments, we used $\mathrm{LSTM}$ s 300 hidden units for $\mathrm{HCLM}$ models. For the cache model, we used cache size 100 in every experiment. All the parameters including character projection parameters are randomly sampled from uniform distribution from $-0.08$ to $0.08$. The initial hidden and memory state of $\mathrm{LSTM}_{\textit{enc}}$ and $\mathrm{LSTM}_{\textit{ctx}}$ are initialized with zero. 
Mini-batches of size 25 are used for PTB experiments and 10 for WikiText-2, due to memory limitations. The sequences were truncated with 35 words. Then the words are decomposed to characters and fed into the model. A Dropout rate of $0.5$ was used for all but the recurrent connections.
}

\paragraph{Learning}
The models were trained with the Adam update rule~\cite{kingma2014adam} with a learning rate of 0.002. The maximum norm of the gradients was clipped at 10.

\paragraph{Evaluation}
We evaluated our models with bits-per-character (bpc) a standard evaluation metric for character-level language models. Following the definition in \newcite{graves2013generating}, bits-per-character is the average value of $-\log_{2} p(w_{t}\mid \boldsymbol{w}_{<t})$ over the whole test set,
\begin{align*}
    \textit{bpc} = -\frac{1}{|\boldsymbol{c}|} \log_{2} p(\boldsymbol{w}) \ignore{p(c_{i}\mid \boldsymbpl{c}_{<i})},
\end{align*}
where $|\boldsymbol{c}|$ is the length of the corpus in characters.
\ignore{
\kk{Notation for character count and word count. Not $N$ for both.}\cd{double check that the $+1$ is correct, since I think we want to count spaces too.} \kk{Yes we need it!}
}

\ignore{

Similarly to perplexity, bits-per-character is also computed from the word-level negative log likelihood. Since the word probability can be written as
\begin{align*}
    p(w) = \prod_{c\in w} p(c),
\end{align*}
the sum of the per-character negative log likelihood is the same as that of words
\begin{align*}
    - \sum_{c} \log_{2}\Pr(c) = -\sum_{w} \log_{2}\Pr(w).
\end{align*}

Thus, the bits-per-character can be computed by dividing the sum of per-word negative log likelihood with the number of characters in dev/test set.}

\subsection{Results}
\paragraph{PTB} Tab.~\ref{tb:result-ptb} summarizes results on the PTB dataset.\footnote{Models designated with a * have more layers and more parameters.} Our baseline $\mathrm{HCLM}$ model achieved 1.276 bpc which is better performance than the $\mathrm{LSTM}$  with Zoneout regularization~\cite{krueger2016zoneout}. And $\mathrm{HCLM}$ with cache outperformed the baseline model with 1.247 bpc and achieved competitive results with state-of-the-art models with regularization on recurrence weights, which was not used in our experiments. \ignore{We note that the sophisticated models such as HyperLSTM~\cite{ha2016hypernetworks} can be augmented with cache components.}

Expressed in terms of per-word perplexity (i.e., rather than normalizing by the length of the corpus in characters, we normalize by words and exponentiate), the test perplexity on $\mathrm{HCLM}$ with cache is 94.79. The performance of the unregularized 2-layer $\mathrm{LSTM}$ with 1000 hidden units on word-level PTB dataset is 114.5 and the same model with dropout achieved 87.0. Considering the fact that our character-level models are dealing with an open vocabulary without unknown tokens, the results are promising.

\begin{table}[ht]
    \begin{center}\small
    \begin{tabular}{l@{\hskip 2pt}r@{\hskip 4pt}r}
    \toprule[0.3ex]
    Method & Dev & Test \\
    \midrule[0.2ex]
    \ignore{
    Norm-stabilized RNN~\cite{krueger2016regularizing} & - & 1.48\\
    }
    CW-RNN~\cite{koutnik2014clockwork} & - & 1.46\\
    HF-MRNN~\cite{mikolov2012subword} & - & 1.41\\
    MI-RNN~\cite{wu2016multiplicative} & - & 1.39\\
    ME $n$-gram~\cite{mikolov2012subword} & - & 1.37\\
    RBN~\cite{cooijmans2016recurrent} & 1.281 & 1.32\\
    Recurrent Dropout~\cite{semeniuta2016recurrent} &  1.338 & 1.301\\
	Zoneout~\cite{krueger2016zoneout} & 1.362 & 1.297\\
	HM-LSTM~\cite{chung2016hierarchical} & - &  1.27\\
	HyperNetwork~\cite{ha2016hypernetworks} & 1.296 & 1.265\\
	LayerNorm HyperNetwork~\cite{ha2016hypernetworks} & 1.281 & 1.250\\
	2-LayerNorm HyperLSTM~\cite{ha2016hypernetworks}* & - & 1.219\\
	2-Layer with New Cell~\cite{zoph2016neural}* & - & 1.214\\
	\midrule
	LSTM (Our Implementation) & 1.369 & 1.331\\
	\ignore{Lily, clm_ptb_8937}
	HCLM \ignore{(d=600)} & 1.308 & 1.276\\
	\textbf{HCLM with Cache \ignore{(d=600)}} & \textbf{1.266} & \textbf{1.247}\\
    \bottomrule[0.3ex]
    \end{tabular}
    \end{center}
        \caption{Results on PTB Corpus (bits-per-character). HCLM augmented with a cache obtains the best results among models which have approximately the same numbers of parameter as single layer $\mathrm{LSTM}$ with 1,000 hidden units. \label{tb:result-ptb}}
        \vskip-1ex
\end{table}

\paragraph{WikiText-2} Tab.~\ref{tb:result-wiki2} summarizes results on the WikiText-2 dataset. Our baseline, $\mathrm{LSTM}$ achieved 1.803 bpc and $\mathrm{HCLM}$ model achieved 1.670 bpc. The $\mathrm{HCLM}$ with cache outperformed the baseline models and achieved 1.500 bpc. The word level perplexity is 227.30, which is quite high compared to the reported word level baseline result 100.9 with $\mathrm{LSTM}$  with ZoneOut and Variational Dropout regularization~\cite{merity2016pointer}. However, the character-level model is dealing with 76,136 types in training set and 5.87\% OOV rate where the word level models only use 33,278 types without OOV in test set. The improvement rate over the $\mathrm{HCLM}$ baseline is 10.2\% which is much higher than the improvement rate obtained in the PTB experiment. 
\begin{table}[ht]
    \begin{center}\small
    \begin{tabular}{l@{\hskip 80pt}r@{\hskip 4pt}r}
    \toprule[0.3ex]
    Method & Dev & Test \\
    \midrule[0.2ex]
    $\mathrm{LSTM}$ & 1.758 & 1.803\\
	HCLM & 1.625 & 1.670\\
	\textbf{HCLM with Cache} & \textbf{1.480} & \textbf{1.500}\\
    \bottomrule[0.3ex]
    \end{tabular}
    \end{center}
        \caption{Results on WikiText-2 Corpus . \label{tb:result-wiki2}}
        \vskip-1ex
\end{table}

\paragraph{Multilingual Wikipedia Corpus (MWC)}
Tab.~\ref{tb:result-mwc} summarizes results on the MWC dataset. Similarly to WikiText-2 experiments, $\mathrm{LSTM}$ is strong baseline. We observe that the cache mechanism improve performance in every languages. In English, $\mathrm{HCLM}$ with cache achieved 1.538 bpc where the baseline is 1.622 bpc. It is 5.2\% improvement. For other languages, the improvement rates were 2.7\%, 3.2\%, 3.7\%, 2.5\%, 4.7\%, 2.7\% in FR, DE, ES, CS, FI, RU respectively. The best improvement rate was obtained in Finnish.

\begin{table*}[ht]
    \begin{center}\small
    \begin{tabular}{lr@{\hskip 4.5pt}r@{\hskip 7pt}r@{\hskip 4.5pt}r@{\hskip 7pt}r@{\hskip 4.5pt}r@{\hskip 7pt}r@{\hskip 4.5pt}r@{\hskip 7pt}r@{\hskip 4.5pt}r@{\hskip 7pt}r@{\hskip 4.5pt}r@{\hskip 7pt}r@{\hskip 4.5pt}r}
    \toprule[0.3ex]
      &
      \multicolumn{2}{c}{EN} &
      \multicolumn{2}{c}{FR} &
      \multicolumn{2}{c}{DE} &
      \multicolumn{2}{c}{ES} &
      \multicolumn{2}{c}{CS} &
      \multicolumn{2}{c}{FI} &
      \multicolumn{2}{c}{RU}\\
    \midrule[0.2ex]
     &
     \multicolumn{1}{c}{dev} & \multicolumn{1}{c}{test} &
     \multicolumn{1}{c}{dev} & \multicolumn{1}{c}{test} &
     \multicolumn{1}{c}{dev} & \multicolumn{1}{c}{test} &
     \multicolumn{1}{c}{dev} & \multicolumn{1}{c}{test} &
     \multicolumn{1}{c}{dev} & \multicolumn{1}{c}{test} &
     \multicolumn{1}{c}{dev} & \multicolumn{1}{c}{test} &
     \multicolumn{1}{c}{dev} & \multicolumn{1}{c}{test}
     \\
    \midrule[0.2ex]
    $\mathrm{LSTM}$ & 1.793 & 1.736 & 1.669 & 1.621 & 1.780 & 1.754 & 1.733 & 1.667 & 2.191 & 2.155 & 1.943 & 1.913 & 1.942 & 1.932\\[0.2ex]
    $\mathrm{HCLM}$ & 1.683 & 1.622 & 1.553 & 1.508 & 1.666 & 1.641 & 1.617 & 1.555 & 2.070 & 2.035 & 1.832 & 1.796 & 1.832 & 1.810\\[0.2ex]
    \textbf{HCLM with Cache} & \textbf{1.591} & \textbf{1.538} & \textbf{1.499} & \textbf{1.467} & \textbf{1.605} & \textbf{1.588} & \textbf{1.548} & \textbf{1.498} & \textbf{2.010} & \textbf{1.984} & \textbf{1.754} & \textbf{1.711} & \textbf{1.777} & \textbf{1.761}\\[0.2ex]
    \bottomrule[0.3ex]
    \end{tabular}
    \end{center}
        \vskip-1.ex
        \caption{Results on MWC Corpus (bits-per-character).\label{tb:result-mwc}}
        \vskip-1.5ex
\end{table*}

\section{Analysis}\label{sec:analysis}
In this section, we analyse the behavior of proposed model qualitatively. To analyse the model, we compute the following posterior probability which tell whether the model used the cache given a word and its preceding context. Let $z_t$ be a random variable that says whether to use the cache or the LM to generate the word at time $t$. We would like to know, given the text $\boldsymbol{w}$, whether the cache was used at time $t$. This can be computed as follows:
\begin{align*}
  p(z_t\mid \boldsymbol{w}) &= \frac{p(z_{t} , w_t \mid \mathbf{h}_t, \text{cache}_{t})}{p(w_t \mid \mathbf{h}_t, \text{cache}_{t})}\\
  &= \frac{(1-\lambda_{t}) p_{\textit{ptr}}(w_t \mid \mathbf{h}_t, \text{cache}_{t})}{p(w_t \mid \mathbf{h}_t, \text{cache}_{t})},
\end{align*}
where $\text{cache}_t$ is the state of the cache at time $t$. We report the average posterior probability of cache generation excluding the first occurrence of $w$, $\overline{p(z \mid w)}$.

\begin{table}[ht]
    \begin{center}\small
    \begin{tabular}{l@{\hskip 25pt}r|l@{\hskip 25pt}r}
    \toprule[0.3ex]
    Word & $\overline{p(z\mid \boldsymbol{w})}\downarrow$ & Word & $\overline{p(z \mid w)}\uparrow$ \\
    \midrule[0.2ex]
. & 0.997 & 300 & 0.000\\
Lesnar & 0.991 & act & 0.001\\
the & 0.988 & however & 0.002\\
NY & 0.985 & 770 & 0.003\\
Gore & 0.977 & put & 0.003\\
Bintulu & 0.976 & sounds & 0.004\\
Nerva & 0.976 & instead & 0.005\\
, & 0.974 & 440 & 0.005\\
UB & 0.972 & similar & 0.006\\
Nero & 0.967 & 27 & 0.009\\
Osbert & 0.967 & help & 0.009\\
Kershaw & 0.962 & few & 0.010\\
Manila & 0.962 & 110 & 0.010\\
Boulter & 0.958 & Jersey & 0.011\\
Stevens & 0.956 & even & 0.011\\
Rifenburg & 0.952 & y & 0.012\\
Arjona & 0.952 & though & 0.012\\
of & 0.945 & becoming & 0.013\\
31B & 0.941 & An & 0.013\\
Olympics & 0.941 & unable & 0.014\\
    \bottomrule[0.3ex]
    \end{tabular}
    \end{center}
        \caption{Word types with the highest/lowest average posterior probability of having been copied from the cache while generating the test set. The probability tells whether the model used the cache given a word and its context. \textbf{Left:} Cache is used for frequent words (\emph{the}, \emph{of}) and proper nouns (\emph{Lesnar}, \emph{Gore}). \textbf{Right:} Character level generation is used for basic words and numbers.}\label{tb:analysis}
        \vskip-1.5ex
\end{table}

Tab.~\ref{tb:analysis} shows the words in the WikiText-2 test set that occur more than 1 time that are most/least likely to be generated from cache and character language model (words that occur only one time cannot be cache-generated). We see that the model uses the cache for proper nouns: \emph{Lesnar}, \emph{Gore}, etc., as well as very frequent words which always stored somewhere in the cache such as single-token punctuation, \emph{the}, and \emph{of}. In contrast, the model uses the language model to generate numbers (which tend not to be repeated): \emph{300}, \emph{770} and basic content words: \emph{sounds}, \emph{however}, \emph{unable}, etc. This pattern is similar to the pattern found in empirical distribution of frequencies of rare words observed in prior wors~\cite{church:1995,church2000empirical}, which suggests our model is learning to use the cache to account for bursts of rare words.

\begin{figure}[htbp]
    \begin{center}
        \includegraphics[width=1.\linewidth]{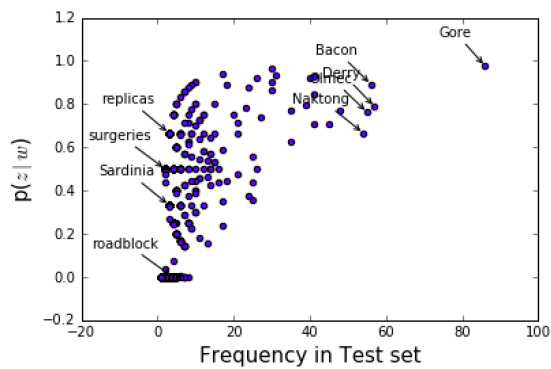}
    \end{center}
     \vskip-1.5ex
    \caption{Average $p(z\mid w)$ of OOV words in test set vs. term frequency in the test set for words not obsered in the training set. The model prefers to copy frequently reused words from cache component, which tend to names (upper right) while character level generation is used for infrequent open class words (bottom left).}
    \label{fig:scatter}
    \vskip-1ex
\end{figure}

To look more closely at rare words, we also investigate how the model handles words that occurred between 2 and 100 times in the test set, but fewer than 5 times in the training set. Fig.~\ref{fig:scatter} is a scatter plot of $\overline{p(z \mid w)}$ vs the empirical frequency in the test set. As expected, more frequently repeated words types are increasingly likely to be drawn from the cache, but less frequent words show a range of cache generation probabilities.

Tab.~\ref{tb:analysis2} shows word types with the highest and lowest average $p(z \mid w)$ that occur fewer than 5 times in the training corpus. The pattern here is similar to the unfiltered list: proper nouns are extremely likely to have been cache-generated, whereas numbers and generic (albeit infrequent) content words are less likely to have been.

\begin{table}[ht]
    \begin{center}\small
    \begin{tabular}{l@{\hskip 4.5pt}r|lr}
    \toprule[0.3ex]
    Word & $\overline{p(z \mid \boldsymbol{w})}\downarrow$ & Word & $\overline{p(z \mid w)}\uparrow$ \\
    \midrule[0.2ex]
    Gore & 0.977 & 770 & 0.003\\
	Nero & 0.967 & 246 & 0.037\\
	Osbert & 0.967 & Lo & 0.074\\
	Kershaw & 0.962 & Pitcher & 0.142\\
	31B & 0.941 & Poets & 0.143\\
	Kirby & 0.935 & popes & 0.143\\
	CR & 0.926 & Yap & 0.143\\
	SM & 0.924 & Piso & 0.143\\
	impedance & 0.923 & consul & 0.143\\
	Blockbuster & 0.900 & heavyweight & 0.143\\
	Superfamily & 0.900 & cheeks & 0.154\\
	Amos & 0.900 & loser & 0.164\\
	Steiner & 0.897 & amphibian & 0.167\\
	Bacon & 0.893 & squads & 0.167\\
	filters & 0.889 & los & 0.167\\
	Lim & 0.889 & Keenan & 0.167\\
	Selfridge & 0.875 & sculptors & 0.167\\
	filter & 0.875 & Gen. & 0.167\\
	Lockport & 0.867 & Kipling & 0.167\\
	Germaniawerft & 0.857 & Tabasco & 0.167\\
	\bottomrule[0.3ex]
    \end{tabular}
    \end{center}
        \caption{Same as Table 7, except filtering for word types that occur fewer than 5 times in the training set. The cache component is used as expected even on rare words: proper nouns are extremely likely to have been cache-generated, whereas numbers and generic content words are less likely to have been; this indicates both the effectiveness of the prior at determining whether to use the cache and the burstiness of proper nouns.\label{tb:analysis2}}
        \vskip-1.5ex
\end{table}

\section{Discussion}
Our results show that the $\mathrm{HCLM}$ outperforms a basic $\mathrm{LSTM}$. With the addition of the caching mechanism, the $\mathrm{HCLM}$ becomes consistently more powerful than both the baseline $\mathrm{HCLM}$ and the $\mathrm{LSTM}$. This is true even on the PTB, which has no rare or OOV words in its test set (because of preprocessing), by caching repetitive common words such as \emph{the}. In true open-vocabulary settings (i.e., WikiText-2 and MWC), the improvements are much more pronounced, as expected. 

\paragraph{Computational complexity.} In comparison with word-level models, our model has to read and generate each word character by character, and it also requires a softmax over the entire memory at every time step. However, the computation is still linear in terms of the length of the sequence, and the softmax over the memory cells and character vocabulary are much smaller than word-level vocabulary. On the other hand, since the recurrent states are updated once per character (rather than per word) in our model, the distribution of operations is quite different. Depending on the hardware support for these operations (repeated updates of recurrent states vs. softmaxes), our model may be faster or slower. However, our model will have fewer parameters than a word-based model since most of the parameters in such models live in the word projection layers, and we use LSTMs in place of these.

\paragraph{Non-English languages.} For non-English languages, the pattern is largely similar for non-English languages. This is not surprising since morphological processes may generate forms that are related to existing forms, but these still have slight variations. Thus, they must be generated by the language model component (rather than from the cache). Still, the cache demonstrates consistent value in these languages.

Finally, our analysis of the cache on English does show that it is being used to model word reuse, particularly of proper names, but also of frequent words. While empirical analysis of rare word distributions predicts that names would be reused, the fact that cache is used to model frequent words suggests that effective models of language should have a means to generate common words as units. Finally, our model disfavors copying numbers from the cache, even when they are available. This suggests that it has learnt that numbers are not generally repeated (in contrast to names).

\ignore{Our results show that the caching mechanism adds considerable power to the $\mathrm{HCLM}$; however, for a fixed number of parameters, the baseline $\mathrm{LSTM}$  outperforms the HCLM. Even if the word is known, it is costly to generate words character by character. Thus, the most effective case will be the language which have long words and high OOVs rate. Fig.\ref{fig:mwc-di scussion} show relationship between average length of words and improvements with cache over baseline $\mathrm{HCLM}$ models. Finnish is the ideal language for the model because it has long words and the highest OOV rate among 7 languages (Table.\ref{tb:uwc}). German also have long words but its OOV rate is actually lower than English.

\begin{figure}[htbp]
    \begin{center}
        \includegraphics[width=0.9\linewidth]{discussion.eps}
    \end{center}
    \caption{Relationship between average word length and performance gain with cache}
    \label{fig:mwc-discussion}
\end{figure}

\ignore{
Fig.~\ref{fig:cache_size} show results on the PTB dataset with different cache size. We found that our model obtained best results with cache size 100 which is the same as the hyperparameter used by \newcite{merity2016pointer}. Ideally, we would like to have large cache to memorize everything mentioned in the past. However, it is easy to imagine that, if there is a large cache, the pointer distribution get flat because of the ambiguous choices in the cache. \kk{Explain the learning challenge we have in the model.}

\kk{Write Analysis, How the cache are used. Try computing posterior.}
}}

\section{Related Work}
Caching language models were proposed to account for burstiness by \newcite{kuhn1990cache}, and recently, this idea has been incorporated to augment neural language models with a caching mechanism~\cite{merity2016pointer,grave2016improving}.

Open vocabulary neural language models have been widely explored~\cite[\emph{inter alia}]{sutskever2011generating,mikolov2012subword,graves2013generating}.\ignore{\cd{please add citations here, make sure to get Sutskever, Mikolov}} Attempts to make them more aware of word-level dynamics, using models similar to our hierarchical formulation, have also been proposed~\cite{chung2016hierarchical}.

The only models that are open vocabulary language modeling together with a caching mechanism are the nonparametric Bayesian language models based on hierarchical Pitman--Yor processes which generate a lexicon of word types using a character model, and then generate a text using these~\cite{teh2006hierarchical,goldwater2009bayesian,chahuneau2013knowledge}. These, however, do not use distributed representations on RNNs to capture long-range dependencies.

\ignore{

\paragraph{Memory Augmented Neural Networks}
In theory, neural networks can model long range dependencies. However, in practice, it is not easy to optimize the model to capture long range dependencies. To explicitly memorize recent history and use them adaptively, ~\newcite{bahdanau2014neural} proposed an attention mechanism. The model store recent history as continuous vectors and use them by computing weighted average of memory vectors. The effectiveness of the attention mechanism was shown in conditional language modeling task. Recently, various types of memory architectures~\cite{graves2014neural,grefenstette2015learning,sukhbaatar2015end,graves2016hybrid} have been investigated. Those models are shown to be effective in tasks which require exact memory such as copying strings. Poninter networks~\cite{vinyals2015pointer} is one of the memory mechanism that is commonly used in language modeling. ~\newcite{merity2016pointer} propose a language model which adaptively modify predicted word distribution with pointer distribution which is computed over recent history.
~\newcite{ling2015finding} proposed another mechanism which store words in memory and, at generation step, the model decide whether the model should copy a word from memory or to generate a word at character-level. They showed its effectiveness in code generation task.
}

\section{Conclusion}
In this paper, we proposed a character-level language model with an adaptive cache which selectively assign word probability from past history or character-level decoding. And we empirically show that our model efficiently model the word sequences and achieved better perplexity in every standard dataset. To further validate the performance of our model on different languages, we collected multilingual wikipedia corpus for 7 typologically diverse languages. We also show that our model performs better than character-level models by modeling burstiness of words in local context.

The model proposed in this paper assumes the observation of word segmentation. Thus, the model is not directly applicable to languages, such as Chinese and Japanese, where word segments are not explicitly observable. We will investigate a model which can marginalise word segmentation as latent variables in the future work.

\section*{Acknowledgements}
We thank the three anonymous reviewers for their valuable feedback. The third author acknowledges the support of the EPSRC and nvidia Corporation.

\bibliography{paper}
\bibliographystyle{acl_natbib}

\newpage
\appendix
\section{Corpus Statistics}\label{sec:appendix-corpus}
\begin{figure*}[ht]
    \begin{center}
        \includegraphics[width=1.0\linewidth]{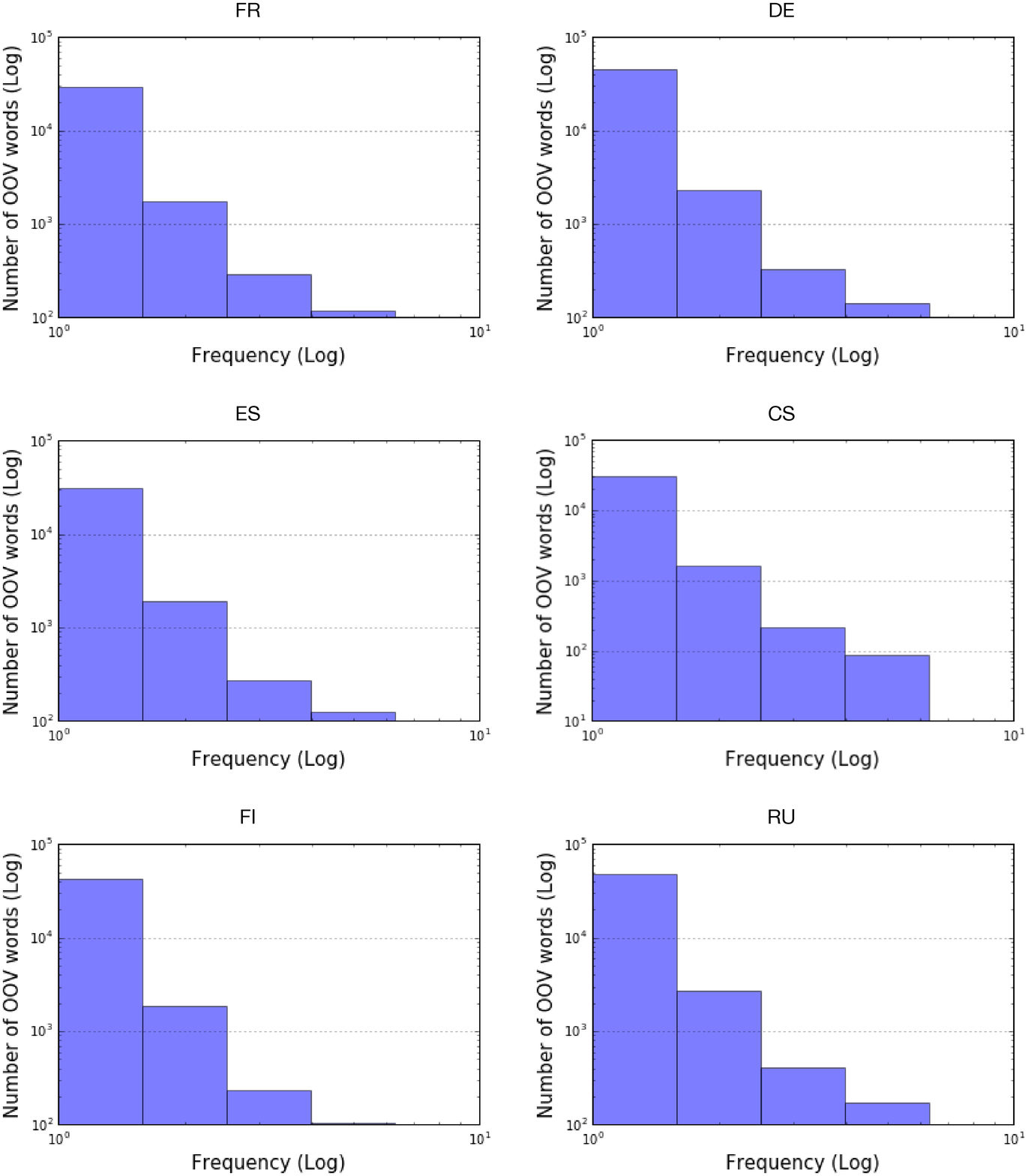}
    \end{center}
    \caption{Histogram of OOV word frequencies in MWC Corpus in different languages.}
    \label{fig:appendix-freq}
\end{figure*}

Fig.~\ref{fig:appendix-freq} show distribution of frequencies of OOV word types in 6 languages.

\end{document}